\documentclass[sigconf]{acmart}
\AtBeginDocument{%
  }

\copyrightyear{2025} 
\acmYear{2025} 
\setcopyright{acmlicensed}\acmConference[WWW Companion '25]{Companion
 Proceedings of the ACM Web Conference 2025}{April 28-May 2, 2025}{Sydney,
 NSW, Australia}
 \acmBooktitle{Companion Proceedings of the ACM Web Conference 2025 (WWW
 Companion '25), April 28-May 2, 2025, Sydney, NSW, Australia}
 \acmDOI{10.1145/3701716.3715229}
 \acmISBN{979-8-4007-1331-6/25/04}

\usepackage{amsmath}
\usepackage{algorithmic}
\usepackage{graphicx}
\usepackage{textcomp}
\usepackage{xcolor}
\usepackage{multicol}
\usepackage{amsthm}
\usepackage[linesnumbered,ruled]{algorithm2e}
\usepackage{caption}
\usepackage{subfigure}
\usepackage{url}
\usepackage{multirow}
\usepackage{float}
\usepackage{enumitem}

\begin{document}

\title{Global Feature Enhancing and Fusion Framework for Strain Gauge Time Series Classification}

\author{Xu Zhang}

\affiliation{%
  \institution{Shanghai Key Laboratory of Data Science, School of Computer Science\\ Fudan University}
  \city{Shanghai}
  \country{China}
}
\email{xuzhang22@m.fudan.edu.cn}

\author{Peng Wang}
\authornote{Peng Wang is the corresponding author.}
\affiliation{%
  \institution{Shanghai Key Laboratory of Data Science, School of Computer Science\\ Fudan University}
  \city{Shanghai}
  \country{China}
}
\email{pengwang5@fudan.edu.cn}

\author{Chen Wang}
\affiliation{%
  \institution{National Engineering Research Center for Big Data Software \\ Tsinghua University}
  \city{Beijing}
  \country{China}
}
\email{wang_chen@mail.tsinghua.edu.cn}

\author{Zhe Xu}
\affiliation{%
  \institution{School of Software \\ Tsinghua University}
  \city{Beijing}
  \country{China}
}
\email{xu_zhe@mail.tsinghua.edu.cn}

\author{Xiaohua Nie}
\affiliation{%
  \institution{National Key Laboratory of Strength and Structural Integrity, Aircraft Strength Research Institute of China}
  \city{Xi'an}
  \country{China}
}
\email{392292400@qq.com}

\author{Wei Wang}
\affiliation{%
  \institution{Shanghai Key Laboratory of Data Science, School of Computer Science\\ Fudan University}
  \city{Shanghai}
  \country{China}
}
\email{weiwang1@fudan.edu.cn}

\renewcommand{\shortauthors}{Xu Zhang et al.}

\begin{abstract}
  Strain Gauge Status (SGS) time series recognition is crucial in the field of intelligent manufacturing based on the Internet of Things, as accurate identification helps timely detection of failed mechanical components, avoiding accidents. 
The loading and unloading sequences generated by strain gauges can be identified through time series classification (TSC) algorithms.
Recently, deep learning models, e.g., convolutional neural networks (CNNs) have shown remarkable success in the TSC task, as they can extract discriminative local features from the subsequences to identify the time series. 
However, we observe that only the local features may not be sufficient for expressing the time series, especially when the local sub-sequences between different time series are very similar, e.g., SGS data of aircraft wings in static strength experiments.
Nevertheless, CNNs suffer from the limitation in extracting global features due to the nature of convolution operations. 
For extracting global features to more comprehensively represent the SGS time series, we propose two insights: (i) Constructing global features through feature engineering. (ii) Learning high-order relationships between local features to capture global features. 
To realize and utilize them, we propose a hypergraph-based global feature learning and fusion framework, which learns and fuses global features for semantic consistency to enhance the representation of SGS time series, thereby improving recognition accuracy. 
Our method designs are validated on industrial SGS and public UCR datasets, showing better generalization for unseen data in SGS recognition. The code is available at the link \url{https://github.com/Meteor-Stars/GFEF}.
\end{abstract}



\maketitle
\section{Introduction}
Strain gauges are widely used in the field of intelligent manufacturing for monitoring strain and deformation, which usually produces large strain gauge data in IoT (Internet of things)-based web server for analysing and ensuring the safety reliability of mechanical components and structures~\cite{DBLP:journals/sensors/LinLTCTW24,DBLP:conf/i2mtc/FarhatYNB24,DBLP:journals/sensors/NiuttaTCP23,DBLP:journals/sensors/EsbeenFVKBGKC22,DBLP:journals/sensors/WernerESTDD22}. 
In our scenario of static strength tests on aircraft wings, the strain data collected during loading and unloading tests are stored in real-time on the web server to monitor Strain Gauges Status (SGS). 
This allows for early warning of mechanical components, e.g., if buckling is detected, the component has the risk of failure under operating conditions and should be replaced promptly.

Currently, business personnel mainly rely on manual classification of SGS based on experience, which is inefficient and costly. 
Moreover, for large mechanical components, the number of strain gauges can reach tens of thousands~\cite{DBLP:journals/sensors/EsbeenFVKBGKC22,DBLP:conf/i2mtc/FarhatYNB24} and this data will continue to climb as the structure become more complex and larger.
Hence, how to automatically extract key features from strain data to accurately classify the SGS for early warning of structural failure is a significant challenge.

\begin{figure}[bt]
\vspace{-0.2cm}
\centerline{\includegraphics[width=1\linewidth]{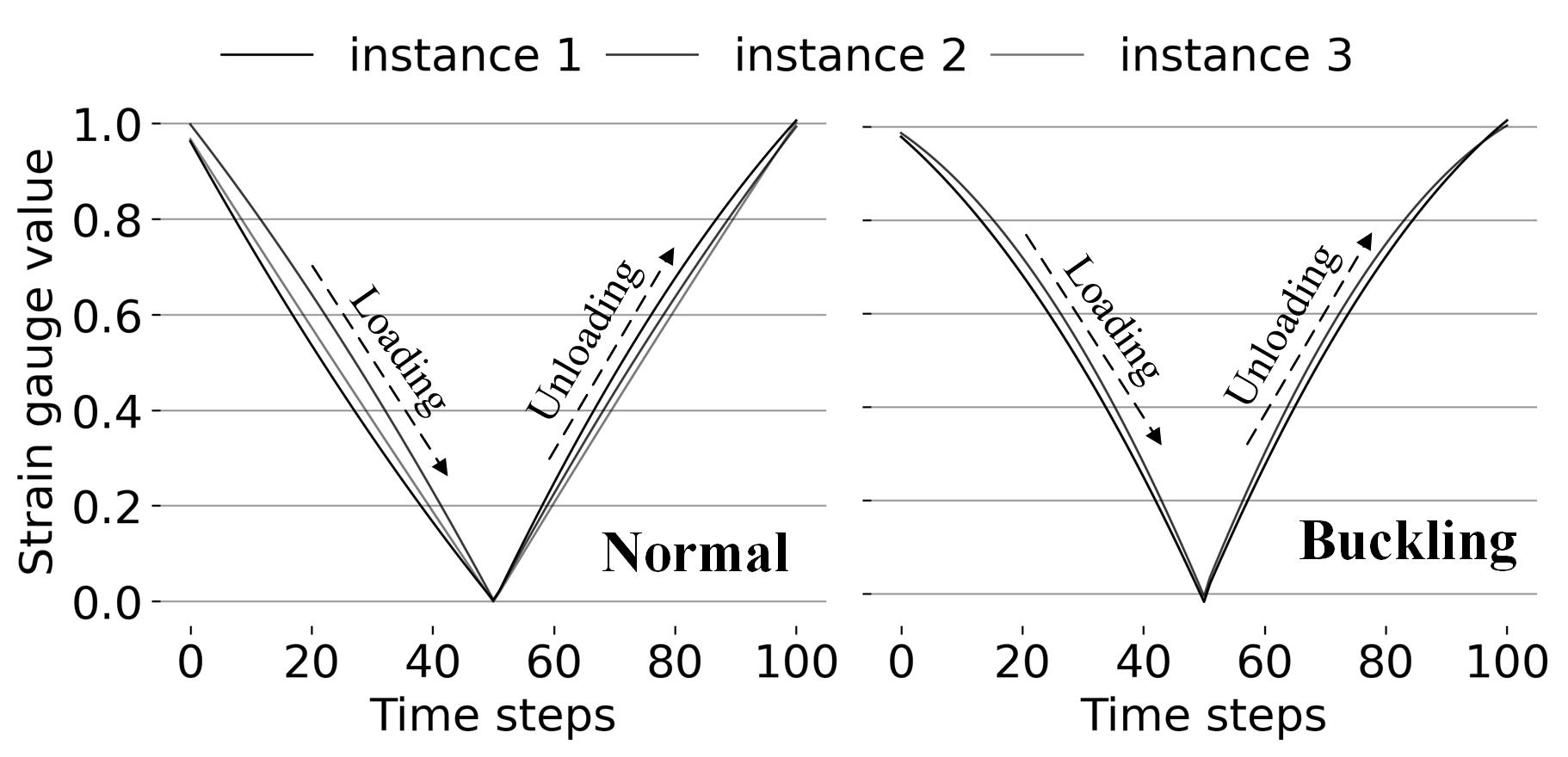} }
\vspace{-0.4cm}
\caption{Time series strain gauge data (after normalization) of aircraft wing obtained from the web server. Experts point out the curvature of buckling curves is generally more pronounced than normal ones.
}
\vspace{-0.3cm}
\label{fig:motivation3}
\end{figure}

During the loading and unloading process of static strength tests, strain gauges generate time-series strain data that change with the load. 
An intuitive idea is to regard the SGS recognition as a Time Series Classification (TSC) task. 
Existing TSC models mainly achieve accurate classification by extracting discriminative local features (local sub-sequences) through shapelets~\cite{DBLP:journals/datamine/YeK11} and one-dimensional convolutional neural networks (1D-CNNs), e.g., ResNet~\cite{fawaz2018data}, InceptionTime~\cite{ismail2020inceptiontime} and Omni-Scale CNN~\cite{tang2020omni}. 
Although 1D-CNNs have achieved remarkable performance in extracting subsequence features for TSC, it suffers
from the limitation in extracting global features due to the nature of convolution operations~\cite{cheng2023formertime,he2022rel}. 
\textbf{Moreover, due to the characteristics (e.g., lower strain value variations) of loading and unloading tests, local subsequences features of strain gauges (e.g., normal and buckling in Figure~\ref{fig:motivation3}) may be very similar, making it difficult for local-feature based models to achieve good accuracy in SGS recognition.} 
This means only the local features may not be sufficient for expressing the time series and the global features that obtained by observing on the whole time series, (e.g., the buckling's curve is more curved on whole than the normal one in Figure~\ref{fig:motivation3}), can be helpful for enhancing the representation and improving TSC accuracy.

For extracting global features for more accurate SGS recognition, our \textbf{one insight} is constructing global features through feature engineering.
Subsequences from different samples may be too similar and have low discriminability, but the whole profile features of the entire sequence can be more distinctive. Hence, \textbf{firstly}, we convert sequences into images to extract profile features to enhance representation.
Due to 2D-CNNs has shown great success in image of handwritten digit classification~\cite{kadam2020cnn} that relies on profile features, we apply it for the image of strain gauge curve to extract the \textbf{profile global feature}. 
Another motivation for using the image is that in business scenarios, experts sometimes can directly judge the image of the strain gauge curve's class visually. In this way, we simulate the visual effects of experts. 
\textbf{Secondly}, we convert expert knowledge into features, e.g., we fit the whole TS with a second-order polynomial based on the Least Squares Method and use the coefficient of the quadratic term as \textbf{expert global feature}, which measures the strain data's overall curvature.

Our \textbf{another insight} is learning high-order relationships between local features to capture global features. 
Rel-CNN~\cite{he2022rel} captures global information by calculating the correlation between local features through differential function while FormerTime~\cite{cheng2023formertime} uses self-attention to model dependence among local features thereby capturing global features. 
However, they all focus solely on pairwise relationships of feature points~\cite{ektefaie2023multimodal}, in contrast, hypergraph~\cite{yadati2019hypergcn,feng2019hypergraph} can consider the high-order correlations and show promise to capture global features from more complex dependence among local features. 
Hence, we propose a hypergraph-based interaction network to realize this insight. 
In this way, we can leverage both local and global features to enhance representations.

Due to different types of global features from two insights are not semantically consistent, directly fusing them (e.g., concatenating and using MLP for fusion) can lead to meaningless intrinsic distribution and thus fail to
maintain the semantic consistency in the feature space, harming the expressive power of learned representations~\cite{wen2023multi,chen2019progressive,xu2019mid}. 
Hence, another key challenge is how to fuse them and utilize their complementarity to improve SGS recognition. 

Fortunately, feature interaction fusion has been demonstrated as an effective way to mitigate the gaps of multi-type features in high dimensional space, realizing feature alignment and re-establishing semantic consistency and enhancing representations~\cite{gong2023skipcrossnets,wang2020learning,wen2023multi,chen2019progressive,xu2019mid}. 
Note that, we previously have mentioned our proposal of a hypergraph interaction network to capture global features. 
Here, we re-use it for interaction fusion of multi-type global features based on its powerful ability of modeling higher-order correlations among feature points, owing to a hyperedge in a hypergraph can connect more than two nodes. 
\textbf{We also observe the high-order interactions resulting in better generalization for unseen data than pairwise interactions, e.g., transformer,} and we will demonstrate this in experiments.  
The reason may be it can quickly establish associations between unseen data and trained data. 
For realizing hypergraph interaction, we transform multi-type global features into nodes and propose constructing cross-feature hyperedges through random walks algorithm~\cite{xia2019random}.

Hypergraph network works through complex information interactions.
However, different global features may contain redundancy and noise. 
If not well processed first, they will be further propagated during the interactions, harming the whole representations. 
To address this, we first design a redundancy filtering module based on learnable binary masks to retain effective information. 
Then, we further propose a data reliability-aware attention mechanism based on a noise perturbation and variance evaluation strategy to reduce the contribution weight of irrelevant feature type.

In general, our contributions are as follows:
\begin{enumerate}[noitemsep, topsep=0pt, wide=\parindent]
\item We design a Global Feature Enhancing and Fusion (GFEF) Framework for SGS recognition, which extracts multi-type global features (profile, expert features, and one learned from local features) and fuses them for enhancing time series representations, utilizing their complementarity to improve SGS recognition accuracy. 
To the best of our knowledge, we are the first study that fuses multi-type features for SGS recognition.  

\item We propose to use learnable binary masks to filter redundancy in features and propose a data reliability-aware attention mechanism to reduce the contribution of irrelevant feature types. They enhance the learned representations by reducing the propagation of redundant noise in the hypergraph interaction process.

\item We have carefully designed the hypergraph interaction module to better extract and fuse global features. The key designs include hypergraph construction based on random walks,  hypergraph attention mechanisms, and dynamic hyperedges. Results show that the high-order interactions from hypergraphs can generate better generalization for unseen data.

\item We perform extensive studies under real industrial scenario about SGS recognition of aircraft wings in static strength experiments, including offline and online studies. Moreover, we also validate our framework on 92 public UCR datasets. 
The results show that our two insights for constructing global features are effective, and our framework can effectively fuse them to enhance feature representation and improve classification accuracy.

\end{enumerate} 
\vspace{-0.1cm}
\section{Related work}
\subsection{Extracting local features for time series classification (TSC)}

Recently, deep learning has been applied to time series classification due to their powerful ability in automatic feature extraction~\cite{Ismail_Muller_2019}, e.g., the 1D-CNNs are extensively exploited to extract local features for TSC~\cite{tang2020rethinking}. 
The typical models are ResNet~\cite{fawaz2018data} and InceptionTime~\cite{ismail2020inceptiontime}.
The latter utilizes convolutional filters of various kernel sizes to extract local features of different scales and achieves promising TSC accuracy. 
However, how to select suitable kernel sizes to cover receptive field remains a challenge. 
To address this, Omni-Scale CNN~\cite{tang2020omni} proposes to use a set of kernel sizes (consisting of multiple prime numbers) to efficiently cover enough receptive field and obtain advanced accuracy. 
Although local feature has been well extracted and shows good effects, it limits accuracy when the local patterns of TS samples are very similar, e.g., in our scenario about SGS recognition. 
In this condition, global features can effectively enhance the discriminability of sample representation.

\vspace{-0.2cm}
\subsection{Extracting global features for time series classification (TSC)}
Existing methods mainly capture global features by learning the dependencies between local features.
Rel-CNN~\cite{he2022rel} extracts global features by calculating the correlations between local features through differential function. 
FormerTime~\cite{cheng2023formertime} combines 1D-CNN and self-attention mechanism to extract global features for TSC. 
However, they all focus solely on pairwise relationships of local features, which may be insufficient for capturing effective global features from the complex dependencies. 
\textbf{In contrast}, we use hypergraph to learn high-order and complex relationships between local features for capturing global features.
Moreover, they only consider extracting global features through raw time series and network design. 
\textbf{In contrast}, we additionally extract global features through feature engineering, i.e., transforming TS into images and expert features as model inputs. 
Finally, we utilize designed hypergraph interaction model for fusing different features, realizing semantic consistency and enhancing whole representations with better generalization.

\vspace{-0.1cm}

\vspace{-0.2cm}
\section{Global Feature Enhancing and Fusion (GFEF) framework for SGS recognition}
We illustrate the proposed framework in Figure~\ref{fig:frame_work}(a). 
Initially, original time series data is transformed into other types, such as image data ($X_{img}$, the curve of $X_{ts}$ in the coordinate axis) and expert features $X_{exp}$ from Table~\ref{tab:expert_feat}. 
For global feature enhancing and fusion based on the hypergraph, we construct node features for different types of inputs. 
Then, we design a feature redundancy filtering and data reliability-aware attention mechanism to reduce the potential redundant and noisy information in the node features, reducing their harm on hypergraph-based representation learning.

Finally, the multi-type node features are sent to the hypergraph interaction layer for fusion and classification. 
Next, we will provide a detailed introduction to each component in the framework.

\begin{table*} [h!] 
    \setlength{\tabcolsep}{1.25pt}
    \centering
    \vspace{-0.2cm}
    \caption{Expert global features for Strain Gauge Status (SGS) recognition. 
    A in feature 3 is any constant.
    $\widehat{L}_{i}$ represents the length of the longest repeated subsequence before the $i$-th time point.
    $\hat{X_{i}}^{(5)}$ denotes the value of the $i$-th time point obtained by fitting a fifth order polynomial. $a_2$ and $a_3$ denote the coefficient of the quadratic and cubic term in second or third order polynomials.}
    \vspace{-0.3cm}
    \label{tab:expert_feat}
    { \small
    \begin{tabular}{c|c|c|c|c|c|c}
        \hline
        \multirow{1}{*}{\shortstack{}}  &  \multicolumn{1}{c|}{Feature 1} &  \multicolumn{1}{c|}{Feature 2}  & \multicolumn{1}{c|}{Feature 3}& \multicolumn{1}{c|}{Feature 4}& \multicolumn{1}{c|}{Feature 5}& \multicolumn{1}{c}{Feature 6}\\
         \midrule[0.5pt]
         \multirow{1}{*}{Formula}  &\shortstack{$max(X_{ts})-min(X_{ts})$} &$ max(\lvert X_{ts} \lvert)$ &$\prod_{i=1}^T (X_i=\text{A}) \in \{0,1\}$ &$\prod_{i=1}^T (\widehat{L}_{i}>\frac{T}{2}) \in \{0,1\}$&$max(\{\lvert X_i-X_{i-1}\lvert\}_{i=2}^T)$&$\frac{2}{T}\sum_{i=1}^{T/2} (X_i-\hat{X_{i}}^{(5)})^2$\\
         \midrule[0.5pt]
        \multirow{1}{*}{\shortstack{}}  &  \multicolumn{1}{c|}{Feature 7} &  \multicolumn{1}{c|}{Feature 8}  & \multicolumn{1}{c|}{Feature 9}& \multicolumn{1}{c|}{Feature 10}& \multicolumn{1}{c|}{Feature 11}& \multicolumn{1}{c}{Feature 12}\\
        \midrule[0.5pt]
           \multirow{1}{*}{Formula}  
          &$\frac{2}{T}\sum_{i=T/2}^{T} (X_i-\hat{X_{i}}^{(5)})^2$ &$\frac{1}{T}\sum_{i=1}^{T} (X_i-\hat{X_{i}}^{(5)})^2$ 
        &$\frac{4}{3T}\sum_{i=T/4}^{T} (X_i-\hat{X_{i}}^{(5)})^2$ 
        &$\frac{4}{3T}\sum_{i=T/4}^{T} (X_i-\hat{X_{i}}^{(2)})^2$ 
          &$Cof_2(\hat{X}^{(2)})=a_2$
          &$\gamma=-a_{2}/3a_{3}-a_{2}^2/3a_{3}^2$
          \\
        \midrule[0.5pt]
    \end{tabular}
    } 
\vspace{-0.3cm}
\end{table*}

\subsection{Feature engineering for global features}
This section introduces profile and expert global features. 
Given the training set $\mathcal{D}={(X_1,Y_1),...,(X_N,Y_N)}$, where $Y \in {1,...,\hat{\text{C}}}$, as a collection of $N$ TS instances with $\hat{\text{C}}$ classes. Each instance $X_{ts}=\{x_1,...,x_T \} \in \mathbb{R}^{1 \times T}$ represents a TS of length $T$, referred to as $X_{ts}$.

To obtain the TS's profile, we convert $X_{ts}$'s curve into 64$\times$64 three-channel image ($X_{img}$), omitting the axes. 
$X_{img}$ includes two forms: one without folding and one folded along the central line of the x-axis. 
Folding helps to make profile distinctions more pronounced, such as for the SGS sequence. 
When the sequence is too long, folding also helps to express profile differences with a shorter image size.
Then, we use four 2D-CNNs to extract profile features, just as extracting the profiles of handwritten digits~\cite{kadam2020cnn}.

For expert global features ($X_{exp}$), we convert expert knowledge into twelve mathematical features by computing on $X_{ts}$, as shown in Table~\ref{tab:expert_feat}. 
For instance, feature 6 to 10 are fitting variances of the high-order polynomials. 
The coefficients of higher-order terms in features 11 and 12 measure the overall curvature of the strain sequence's curve, which is beneficial for the recognition of buckling status. 
These features are obtained by observing the whole sequence of $X_{ts}$ and representing the global information of the strain TS.
We use a simple linear layer to project the twelve expert features into high-dimensional space for learning.

\subsection{Multi-types feature encoder with lower noisy information}
In this section, we introduce how to construct high-reliability inputs, thereby reducing the propagation of redundant noise during hypergraph interaction.
\vspace{-0.1cm}
\subsubsection{\textbf{Node feature construction} }
As discussed above, lightweight 2D-CNNs and simple linear layer are used to extract features from $X_{img}$ and $X_{exp}$. 
For original time series $X_{ts}$, we use OSCNN~\cite{tang2020omni} for feature extraction.
Then, we simply divide each feature into $P$ non-overlapping patches with length $L$ as nodes and use a linear layer to project them into high-dimensional space. 
The nodes feature of $X_{ts}$, $X_{img}$ and $X_{exp}$ are denoted as $Z_{ts}$, $Z_{img}$ and $Z_{exp}$, each of them $\in \mathbb{R}^{P \times d}$ where $P$ is the number of nodes and $d$ is the dimension of feature space.

\vspace{-0.1cm}
\subsubsection{\textbf{Feature redundancy filtering }}
\label{sec:Para_Bern}

As stated in~\cite{wang2023reducing,wang2020orthogonal}, CNNs have significant redundancy between filters and feature channels, harming feature representations. 
For instance, redundancy may affect the model's utilization of truly effective local features (we will further discuss this in Figure~\ref{fig:FRF_heat_map} of the experiment section).
Hence, we propose a feature redundancy filtering module to address this issue. 
For simplicity, we take the example of $Z_{ts}$ to explain the redundancy filtering process.

Specifically, we decouple the feature space of each node into two parts: class-relevant and class-redundant features and our goal is to retain the former and drop the latter in the end-to-end training framework, as shown in Figure~\ref{fig:frame_work}(b).
Specifically, the probabilities of $i$-th ($i \leq d$) feature point of each node are class-redundant or class-relevant ones are defined as $c_{i}=\{ c^0_{i},c^1_{i}\}$ and $c^0_{i}+c^1_{i}=1$, which follows a standard Bernoulli distribution. 
Specifically, we use a classifier to predict the probability based on $Z_{ts}$:
\vspace{-0.1cm}
 \begin{equation}
\label{equ:features_e}
c=Permute(Z_{ts})\times \Theta_1+b_1
\vspace{-0.1cm}
\end{equation}
where $\Theta_1 \in \mathbb{R}^{P \times 2}$ and \textbf{b}$_1$ are learnable parameters. 
$Permute$ denotes permute the last two dimensions.
Now, we can sample the discrete values $z_{i} \in \{0,1\}$ (binary masks) based on the probability $c_{i}$. 
If $c_{i}<0.5$, $z_i=0$ and this denotes $i$-th feature point of nodes in $Z_{ts}$ is redundant and will be discarded, and vice versa.

However, the discrete sampling is non-differentiable.  
To address this, we first use the Gumbel-Max trick~\cite{DBLP:conf/iclr/JangGP17,DBLP:conf/iclr/MaddisonMT17} to make the sampling process of the discrete variable $z_{i}$ differentiable:
\vspace{-0.1cm}
\begin{equation}
\vspace{-0.1cm}
\label{equ:final_scale_factors}
    z_{i}=\mathop{\arg\max}_{a \in \{0,1\}}\text{log} c^a_{i}+g^a_{i}
\end{equation}
where $g^0$ and $g^1$ draw from a standard Gumbel distribution, which is sampled by drawing $u\sim$ Uniform(0,1) and computing $g=-\text{log}(-\text{log}u)$. 

To make $\mathop{\arg\max}$ differentiable, we further substitute it with the Gumbel-Softmax reparameterization trick~\cite{DBLP:conf/iclr/JangGP17,DBLP:conf/iclr/MaddisonMT17} as:
\vspace{-0.2cm}
 \begin{equation}
 \vspace{-0.1cm}
 \label{equ:learn_connection}
  z_{i} = \sum_{k=0}^1 \frac{exp((\text{log} c^a_{i}+g^a_{i})/\tau)}{\sum_{a\in \{0,1\}} exp((\text{log} c^a_{i}+g^a_{i})/\tau) }\times k
\vspace{-0.1cm}
\end{equation}
where $\tau$ is the temperature parameter to control the Gumbel-Softmax distribution's smoothness.

Now, we can perform unsupervised binary classification learning through standard gradient descent algorithms and the model automatically decides whether to retain the $i$-th feature point of nodes in $Z_{ts}$ through the learned binary masks $z$.

\subsubsection{\textbf{Data reliability-aware attention mechanism}}
In addition to finely filtering redundancy at the node level, we further assign attention weights to different feature types (e.g., $Z_{img}$) to coarsely reduce noise information.
Intuitively, the effectiveness of different types of features for classification varies dynamically. 
Taking image data as an example, when the profile features of samples from different classes are quite similar, the model should automatically assign a lower weight to image data and rely more on other features, making profile features play a supportive rather than a leading role. 

To realize this, we first assign data-specific intermediate classifiers $CL_{ts}$, $CL_{img}$ and $CL_{exp}$ for different types of features. 
For enabling them to measure data quality, we optimize these classifiers using cross-entropy loss. 
Meanwhile, we introduce Jensen-Shannon divergence~\cite{menendez1997jensen} constraint between them and the final classifier, using it to further guide and enhance early classifiers' abilities.

Intuitively, the classifier's confidence can measure the 
importance and reliability of different feature types for the downstream task.
However, as stated in ~\cite{moon2020confidence,van2020uncertainty}, the classifier usually leads to over-confidence, especially for erroneous prediction. 
Hence, directly using classifier's confidence to measure the reliability may not be reliable. 
To address this, we propose to measure the reliability of the features by the classifier's fluctuation in response to noise perturbations.
Intuitively, if the feature type is effective and reliable for classification, it should have strong class distinguishability and be relatively robust under noise perturbations, i.e., the classifier's decision shouldn't be easily changed due to the noise.

\begin{figure*}[bt]
\centerline{\includegraphics[width=1\linewidth]{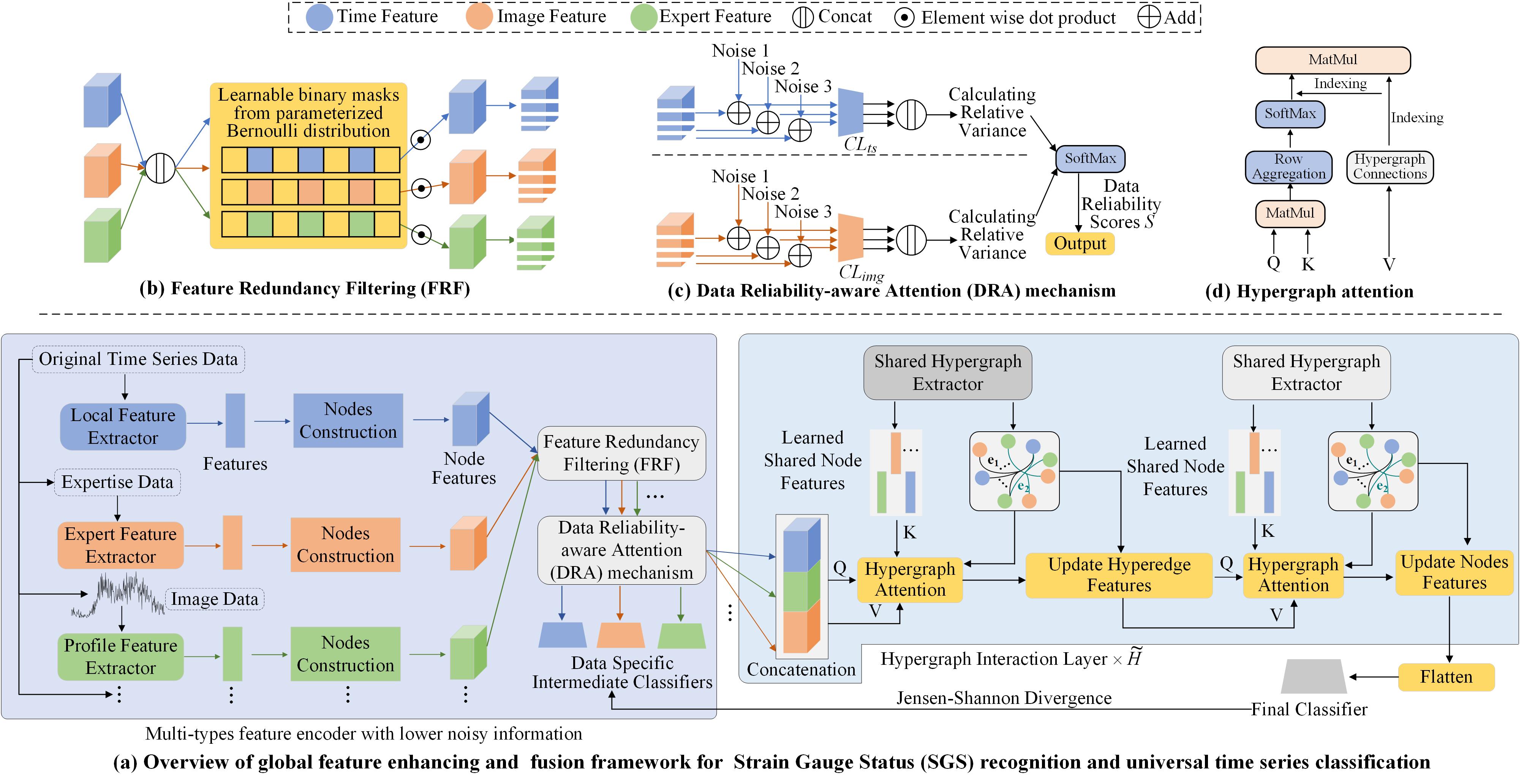} }
\vspace{-0.4cm}
\caption{The proposed Global Feature Enhancing and interaction Fusion (GFEF) framework with its components.}
\label{fig:model}
\vspace{-0.4cm}
\label{fig:frame_work}
\end{figure*}

To realize our idea for measuring the reliability of features, we calculate the relative variance among classes after adding noisy perturbations and apply $softmax(\cdot)$ to obtain the reliability score $\mathcal{S}$ for different types of features, as shown in Figure~\ref{fig:frame_work}(c). 
\textbf{If the feature exhibits greater fluctuation in response to perturbations, it is less important and susceptible to noise, denoting lower reliability.} 
In hypergraph interactions, the model should assign a lower weight to this feature (e.g., $Z_{img}=Z_{img}\cdot\mathcal{S}_{img}$) to reduce the negative impacts of noise information on the feature interaction. 
Taking the calculation of feature reliability scores for $Z_{ts}$ and $Z_{img}$ for instance, the algorithm pseudocode is shown in Algorithm \ref{alg:algorithm1}.
\vspace{-0.2cm}
\begin{algorithm}[h]
\caption{Data reliability-aware attention scores.}
	\label{alg:algorithm1}
	
	\KwIn{
 \begin{enumerate}
 \item Features $Z_{ts}$ and $Z_{img}$. 
 \item Classifiers $CL_{ts}$ and $CL_{img}$. Their respective outputs are initialized as empty sets $O_{ts}$ and $O_{img}$.
 \item Standard Gaussian noise function $\mathcal{G}(\cdot)$ and noise perturbation levels \textit{noise\_levels} $\in \{1,2,3\}$. 
 \item Variance function \textit{Var}($\cdot$) and some math functions \textit{max}($\cdot$), \textit{min}($\cdot$), \textit{log}($\cdot$), \textit{concat}($\cdot$) and \textit{softmax}($\cdot$).
 \end{enumerate}}
        \KwOut{Reliability scores for different types of data $\mathcal{S}$.}
	\BlankLine
            $O_{ts} \Leftarrow \emptyset_{ts}$, $O_{img} \Leftarrow \emptyset_{img}$\\
		\For{$j=1$ to $max(\textit{noise\_lvels})$}{
 $O_{ts} \Leftarrow O_{ts} \cup {CL_{ts}}\left( Z_{ts}\cdot( \mathcal{G}(Z_{ts})+1)\cdot j \right)$\\
 $O_{img} \Leftarrow O_{img} \cup {CL_{img}}\left( Z_{img}\cdot( \mathcal{G}(Z_{img})+1)\cdot j \right)$\\}
Var$_{ts}$=\textit{Var}($O_{ts}$), Var$_{img}$=\textit{Var}($O_{img}$)\\
Relative\_Var$_{ts}$=\textit{max}(Var$_{ts}$)-\textit{min}(Var$_{ts}$)\\
Relative\_Var$_{img}$=\textit{max}(Var$_{img}$)-\textit{min}(Var$_{img}$)\\
\Return \textbf{1}-\textit{softmax}$\left(\text{Relative}\_Var_{ts},\text{Relative}\_Var_{img}\right)$
\end{algorithm}

\vspace{-0.1cm}
\subsection{Global feature learning and interaction fusion}
The hypergraph interaction network serves two purposes. (i) is global feature learning, i.e., learning high-order relationships between
local features to capture global features. 
(ii) is the interaction fusion of global features to mitigate the semantic gaps among them, realizing feature alignment and re-establishing semantic consistency for enhancing representations~\cite{gong2023skipcrossnets,wang2020learning,wen2023multi,chen2019progressive,xu2019mid}. 

\subsubsection{\textbf{Random walk hypergraph construction (RWHC)}}
The degree for all edges is 2 in the normal graph, i.e., pairwise connections. 
In contrast, a hypergraph can encode high-order data correlation using its degree-free hyperedges~\cite{feng2019hypergraph}. 
Hence, we use hypergraphs to capture complex dependence among different feature nodes, further learning and fusing global features.
We define the hypergraph as $H=(V,\mathcal{E})$ where $V$ is a set of nodes and a hyperedge $e \in \mathcal{E}$ is a non-empty subset of $V$. 
Existing methods perform K-Nearest Neighbor (KNN) searches for each node, treating the top-$K$ nearest neighbors along with the original node as a hyperedge~\cite{DBLP:conf/ijcai/JiangWFCG19,feng2019hypergraph}. 

This may not be suitable for our scenario with multi-type node features, e.g., 
if KNN is used, the hyperedge $e_i \in e_{img}$ may only contain nodes of $Z_{img}$, without connection with nodes in $Z_{ts}$ and $Z_{exp}$ (we have demonstrated this in Figure~\ref{fig:hyperG_res} of the appendix). 
This limits the utilization of inherent correlations of cross-type nodes to enhance representations. 
To address this, we relax the similarity constraints and propose constructing hyperedges through the random walk algorithm, which is popularly used for finding important nodes in the graph~\cite{DBLP:journals/cn/BrinP12,rosvall2008maps,perozzi2014deepwalk}.
Random walk makes hyperedges retain the original important nodes while ensuring connections across other feature types. 
For utilizing random walk, we first learn a shared connection matrix $\mathcal{C}$ with dimension of $\hat{P}\times\hat{P}$, where $\hat{P}$ denotes total number of nodes for all features. 
$\mathcal{C}$ is learned through Parameterized Bernoulli distribution (predicting connections between any two nodes), which follows the technique described in section~\ref{sec:Para_Bern}. 
Besides, to save on computation and help the model converge better, we initialize learnable embeddings $E$ for all nodes and use one MLP with three 1D-CNNs to extract feature $\mathcal{M} \in \mathbb {R}^{d \times \hat{P}}$ from $E$ to learn the connection matrix, instead of using original node feature $Z$ to learn it. 
After obtaining the connection matrix $\mathcal{C}$, the process of random walk with restart probability $\alpha \in [0,1]$ is:
\vspace{-0.2cm}
\begin{equation}
\vspace{-0.2cm}
\label{equ:random_walk}
    \widehat{\mathcal{C}}=\sum_{k=0}^c \alpha(1-\alpha)^{k}D^{-1}\mathcal{C}
\end{equation}
where $D^{-1}\mathcal{C}$ is state transition matrix.
$D^{-1}=diag(\mathcal{C})$ is the out-degree diagonal matrix and $c$ denotes the number of random walks.

After random walks, the $i$-th row of the matrix $\widehat{\mathcal{C}}$ records the relevance scores (probabilities) of other nodes with respect to the $i$-th node. 
We take the top-$K$ nodes along with the $i$-th node to form a hyperedge and we can obtain $\hat{P}$ hyperedges. 
We further illustrate the process of hypergraph construction in Figure~\ref{fig:hypg_learn} of the appendix due to limited space.

\subsubsection{\textbf{Hypergraph attention based information propagation}}
\textbf{Hypergraph attention}.
The hypergraph attention is designed to learn the attention weights between nodes and hyperedges. This is because different nodes have varying degrees of importance for the corresponding hyperedges, and vice versa. 

We first measure the influence of the $i$-th node on other nodes through dot product, then aggregate the values of each row to obtain the total influence of the $i$-th node, and finally obtain the importance score through the \textit{softmax} function, which can be formulated as:
\begin{equation}
\vspace{-0.1cm}
\label{equ:attn1}
    Q^v,\mathcal{X}^v=\left( Z_{ts} \parallel Z_{img} \parallel Z_{exp} \right) \Theta_2 +b_2
\end{equation}

\begin{equation}
\label{equ:attn2}
    Att^v= \textit{softmax}(Row\_Agg(Q^v \odot \mathcal{M}^v)) 
\end{equation}
where $\parallel$ denotes concatenation. 
$Row\_Agg$ denotes sum aggregation at row-wise and $\odot$ denotes dot product operation. $Att \in \mathbb{R}^{1\times\hat{P}}$ is importance scores of each node. 

Note that, the hypergraph is constructed using shared node features $\mathcal{M}$. 
Hence, we use shared $\mathcal{M}$ as Key to dot-multiply with private $Q^v$ of each sample, 
thereby enriching gradient flow on hypergraph learning for better optimization. 
The illustration of the proposed attention mechanism is shown in Figure~\ref{fig:frame_work}(d).

\textbf{Updating hyperedges based on ndoes}. Following UniGNN~\cite{DBLP:conf/ijcai/HuangY21}, the propagation of nodes for updating hyperedges is:
\begin{equation}
\label{equ:propagation_nodes_to_edges}
    \mathcal{X}_i^e= \textit{Norm} \left( W \left(\sum_{k \in e_i} Att^v[k]\mathcal{X}_i^v[k,:]\right) \right)
\end{equation}
where $W$ is learnable linear transform, $\mathcal{X}_i^e$ denotes $i$-th hyperedge feature and $k$ denotes node index. \textit{Norm} is Layer Normalization.

\textbf{Updating nodes based on dynamic hyperedges}. 
After updating the hyperedge features, it intuitively follows that the original hyperedge structure should also be updated accordingly. 
However, existing methods~\cite{DBLP:conf/ijcai/HuangY21,DBLP:conf/ijcai/JiangWFCG19,feng2019hypergraph} all use the initial hyperedge structure before updating, which may limit the model's representation learning ability. 
We will demonstrate this point in the ablation study. 
To address this, we learn a new hyperedge structure $\widehat{\mathcal{E}}$ for the updated hyperedges based on random walks and use it to update the node features. 
With residual connection $\mathcal{X}^v$, the process of updating nodes based on new hyperedges $\widehat{\mathcal{E}}$ and attention scores is:
\vspace{-0.1cm}
\begin{equation}
\vspace{-0.1cm}
\label{equ:propagation_nodes_to_edges}
    \mathcal{\widehat{X}}_{i}^{v}=\textit{Norm} \left( W \left(\mathcal{X}^v+ \sum_{k \in v_e} Att^e[k]\mathcal{X}^e[k,:] \right)\right)
\vspace{-0.1cm}
\end{equation}
where $\mathcal{\widehat{X}}_i^v$ denotes $i$-th node feature and $k$ denotes hyperedge index. 
$v_e$ represents the set of hyperedge indices that the $i$-th node belongs.

Finally, the node features after hypergraph interaction fusion are flattened and sent to the classifier $CL_{final}$ for predicting labels. 
The final loss function includes four cross entropy losses of $CL_{ts}$,  $CL_{img}$, $CL_{exp}$, $CL_{final}$ and one Jensen-Shannon divergence loss between $CL_{final}$ and $CL_{ts}$,  $CL_{img}$, $CL_{exp}$. We sum all the losses for end-to-end training. 
The whole hypergraph interaction framework and the components are shown in Figure~\ref{fig:frame_work}(a)-(d).

\section{Experiments and results}
\subsection{Experimental settings}
\vspace{-0.075cm}
\subsubsection{\textbf{Industrial and public datasets}}
We have collected the industrial dataset from the scenario about SGS recognition of aircraft wings in four static strength experiments. 
The four experimental groups produce 500, 9926, 1747, and 1742 samples respectively, with each sample's time series length being 101, denoting the strain values from the loading and unloading process. 
For more comprehensive evaluations, 
each group of data takes turns serving as the training set, and is evaluated on the other three group's data. 
The average metrics of three evaluations are then taken as performance for that dataset. 
\textbf{This approach allows us to evaluate the algorithm's robustness and generalizability across different experiments.}

\textbf{92 UCR datasets from the UCR archive\footnote{ \url{https://www.timeseriesclassification.com/}}.} They are widely used for evaluating TSC methods~\cite{tang2020omni,ismail2020inceptiontime,wang2017time}.

In these original datasets, the training and testing sets have been well processed. We do not take any processing for these datasets for a fair comparison.

\vspace{-0.1cm}
\subsubsection{\textbf{Baselines and metrics}}

\textbf{TSC baselines.} Six advanced TSC baselines are selected for comparing with our method, including InceptionTime~\cite{ismail2020inceptiontime}, OS-CNN~\cite{tang2020omni}, FCNet~\cite{wang2017time}, ResNet~\cite{wang2017time}, FormerTime~\cite{cheng2023formertime} and MultiRocket~\cite{41tan2022multirocket}. 
We compare with them to illustrate that our method can better learn high-order relationships between
local features to capture global features, thereby improving accuracy on single time series input.

\textbf{Interaction fusion baselines.} Five advanced interaction fusion baselines in the TS field are used for comparison, including MLP fusion, CNN fusion, WideDeep fusion~\cite{cheng2016wide}, transformer fusion~\cite{vaswani2017attention}, and adaptive fusion~\cite{wang2022adaptive}. 
\textbf{We compare with them to illustrate that our method can better realize feature alignment and re-establish semantic consistency, thereby enhancing representations and improving TSC accuracy through their complementarity.} The accuracy, F1 score, and precision are used to evaluate all methods, which are widely used in TSC literatures~\cite{ismail2020inceptiontime,cheng2023formertime,tang2020omni}.

\vspace{-0.1cm}
\subsubsection{\textbf{Implementation details.}}
All methods share the same basic settings (batch size, learning rate, etc.) for fair comparison. 
The batch size is set according to the scale of the dataset, the learning rate is fixed at 1e-3, and all methods are trained for 500 epochs. 
The patch length $L$ and stride for constructing nodes are fixed at 8.
The number of nodes $P$ is fixed at 16 for each feature. 
The top-$K$ for selecting nodes to construct hyperedges is fixed at $K=12$. 
The hidden size and layer number $\widetilde{H}$ in the model are fixed at 128 and 1. 
$c$ used in Eq.~\ref{equ:random_walk} for random walk is 1.
Global feature $X_{img}$ includes two forms: one without folding and one folded along the central line of the x-axis. 
For each experiment, we independently run four times with four different seeds, and the average metrics and standard deviations are reported. Experiments are conducted on NVIDIA GeForce RTX 3090 GPU on PyTorch. More experimental results and analysis are shown in the appendix.

\begin{table*}[h]
 \vspace{-0.15cm}
    \setlength{\tabcolsep}{4.5pt}
    \centering
    \caption{Offline strain gauge status (SGS) recognition on four groups of static strength experiments of aircraft wings.}
    \vspace{-0.3cm}
    \label{tab:metric_TSC_SGS}
    {\footnotesize
    \begin{tabular}{c|c|ccc|ccc}
        \hline
        \multirow{2}{*}{\shortstack{}} &\multirow{2}{*}{Methods} &  \multicolumn{3}{c|}{Experimental Group 1}  &  \multicolumn{3}{c}{Experimental Group 2} \\
         & & Average Accuracy & Average F1 Score &Average Precision & Average Accuracy & Average F1 Score &Average Precision \\ 
         \midrule[0.5pt]
         \multirow{6}{*}{\shortstack{Strain gauge datasets \\(only TS input)}} &FCNet~\cite{wang2017time} &86.9$\pm$0.001&85.992$\pm$0.0015&86.533$\pm$0.0025&92.925$\pm$0.0013&92.575$\pm$0.0018&92.917$\pm$0.0014\\
         &ResNet~\cite{wang2017time}&87.533$\pm$0.0012&86.842$\pm$0.0013&87.533$\pm$0.0022&93.133$\pm$0.0019&92.783$\pm$0.0017&93.092$\pm$0.0032\\
         
        &InceptionTime~\cite{ismail2020inceptiontime} &87.833$\pm$0.002&87.175$\pm$0.0025&87.692$\pm$0.0053&93.292$\pm$0.0016&92.942$\pm$0.0014&93.017$\pm$0.0023\\

         &OSCNN~\cite{tang2020omni} &87.033$\pm$0.0008&86.192$\pm$0.0016&86.6$\pm$0.0024&92.667$\pm$0.001&92.275$\pm$0.0009&92.475$\pm$0.0012\\

    &FormerTime~\cite{cheng2023formertime} &85.642$\pm$0.0016&84.175$\pm$0.0042&84.8$\pm$0.0031&92.608$\pm$0.001&92.208$\pm$0.0012&92.475$\pm$0.001\\
    &MultiRocket~\cite{41tan2022multirocket}&86.533$\pm$0.0009&85.917$\pm$0.001&86.583$\pm$0.0012&90.692$\pm$0.0025&90.275$\pm$0.0025&90.483$\pm$0.0026\\
    &Ours &\textbf{88.017$\pm$0.0019}&\textbf{87.308$\pm$0.0021}&\textbf{87.842$\pm$0.0041}&\textbf{93.517$\pm$0.002}&\textbf{93.142$\pm$0.0023}&\textbf{93.5$\pm$0.0018}\\
    
    \hline
    \hline
    \multirow{2}{*}{\shortstack{}} &\multirow{2}{*}{Methods} &  \multicolumn{3}{c|}{Experimental Group 1}  &  \multicolumn{3}{c}{Experimental Group 2} \\
         & & Average Accuracy & Average F1 Score &Average Precision & Average Accuracy & Average F1 Score &Average Precision \\ 
         \midrule[0.5pt]
    \multirow{6}{*}{\shortstack{Strain gauge datasets \\(Fusing TS, Image \\and Expert features)}} 
        &MLP Fusion&90.35$\pm$0.0025&89.383$\pm$0.0037&89.817$\pm$0.0046&93.308$\pm$0.0028&93.142$\pm$0.0039&93.225$\pm$0.0048\\
        &CNN Fusion &91.892$\pm$0.0068&91.542$\pm$0.0074&92.067$\pm$0.0068&96.333$\pm$0.0028&96.292$\pm$0.0029&96.3$\pm$0.0028\\
        &WideDeep Fusion~\cite{cheng2016wide} &88.275$\pm$0.0092&87.425$\pm$0.0085&87.933$\pm$0.0076&93.358$\pm$0.0022&93.15$\pm$0.0035&93.183$\pm$0.0031\\
        &Transformer Fusion~\cite{vaswani2017attention}  &90.625$\pm$0.0022&89.575$\pm$0.0043&90.292$\pm$0.0017&97.025$\pm$0.0018&96.942$\pm$0.0019&96.992$\pm$0.0018\\
         &Adaptive Fusion~\cite{wang2022adaptive} &92.692$\pm$0.0039&92.292$\pm$0.005&92.8$\pm$0.0045&95.725$\pm$0.0165&95.592$\pm$0.0173&95.575$\pm$0.0176\\

    &Ours &\textbf{92.875$\pm$0.0014}&\textbf{92.492$\pm$0.0011}&\textbf{92.967$\pm$0.0016}&\textbf{97.517$\pm$0.0012}&\textbf{97.5$\pm$0.0011}&\textbf{97.525$\pm$0.001}\\
     \hline
     \hline
        \hline
        \multirow{2}{*}{\shortstack{}} &\multirow{2}{*}{Methods} &  \multicolumn{3}{c|}{Experimental Group 3}  &  \multicolumn{3}{c}{Experimental Group 4} \\
         & & Average Accuracy & Average F1 Score &Average Precision & Average Accuracy & Average F1 Score &Average Precision \\ 
         \midrule[0.5pt]
         \multirow{6}{*}{\shortstack{Strain gauge datasets \\(only TS input)}} &FCNet~\cite{wang2017time} &88.8$\pm$0.0027&88.358$\pm$0.0031&89.225$\pm$0.0043&87.325$\pm$0.0044&86.375$\pm$0.0058&86.983$\pm$0.006\\
         &ResNet~\cite{wang2017time}&88.217$\pm$0.0079&88.267$\pm$0.0063&89.358$\pm$0.0059&87.408$\pm$0.0025&86.508$\pm$0.0034&86.992$\pm$0.0026\\
        &InceptionTime~\cite{ismail2020inceptiontime} &89.017$\pm$0.0017&89.108$\pm$0.0016&89.958$\pm$0.0019&87.925$\pm$0.0017&87.308$\pm$0.0019&87.65$\pm$0.0014\\

         &OSCNN~\cite{tang2020omni} &89.375$\pm$0.0033&89.117$\pm$0.003&89.692$\pm$0.002&87.667$\pm$0.0024&86.95$\pm$0.0041&87.292$\pm$0.004\\

    &FormerTime~\cite{cheng2023formertime} &88.492$\pm$0.0023&88.467$\pm$0.0028&89.2$\pm$0.0029&87.467$\pm$0.0006&86.708$\pm$0.0007&87.658$\pm$0.0025\\
    &MultiRocket~\cite{41tan2022multirocket}&85.117$\pm$0.0027&85.675$\pm$0.0025&87.825$\pm$0.0023&85.675$\pm$0.0014&84.775$\pm$0.0014&86.175$\pm$0.0009\\
    &Ours &\textbf{89.642$\pm$0.0023}&\textbf{89.5$\pm$0.0025}&\textbf{90.2$\pm$0.0035}&\textbf{88.45$\pm$0.0011}&\textbf{87.875$\pm$0.0008}&\textbf{88.35$\pm$0.0033}\\
    
    \hline
    \hline
    \multirow{2}{*}{\shortstack{}} &\multirow{2}{*}{Methods} &  \multicolumn{3}{c|}{Experimental Group 3}  &  \multicolumn{3}{c}{Experimental Group 4} \\
         & & Average Accuracy & Average F1 Score &Average Precision & Average Accuracy & Average F1 Score &Average Precision \\ 
         \midrule[0.5pt]
    \multirow{6}{*}{\shortstack{Strain gauge datasets \\(Fusing TS, Image \\and Expert features)}} 
        &MLP Fusion&90.608$\pm$0.0032&90.383$\pm$0.0032&90.858$\pm$0.0028&89.683$\pm$0.0108&89.467$\pm$0.0121&89.675$\pm$0.0116\\
        &CNN Fusion&90.375$\pm$0.0043&90.05$\pm$0.0044&90.592$\pm$0.0045&93.775$\pm$0.0052&93.708$\pm$0.0047&93.958$\pm$0.0048\\
    &WideDeep Fusion~\cite{cheng2016wide} &90.433$\pm$0.0025&90.333$\pm$0.0028&91.067$\pm$0.0041&90.133$\pm$0.0063&89.9$\pm$0.007&90.075$\pm$0.0066\\
&Transformer Fusion~\cite{vaswani2017attention}  &89.858$\pm$0.0022&89.783$\pm$0.0015&90.5$\pm$0.0018&93.4$\pm$0.0047&93.35$\pm$0.0041&93.683$\pm$0.0045\\
         &Adaptive Fusion~\cite{wang2022adaptive} &88.858$\pm$0.021&88.0$\pm$0.0303&88.342$\pm$0.0334&93.383$\pm$0.0092&93.275$\pm$0.0096&93.392$\pm$0.0103\\

    &Ours &\textbf{95.1$\pm$0.002}&\textbf{94.942$\pm$0.0028}&\textbf{95.258$\pm$0.002}&\textbf{94.967$\pm$0.0017}&\textbf{94.908$\pm$0.0013}&\textbf{95.008$\pm$0.0016}\\
    
        \hline
    \end{tabular}}
\vspace{-0.1cm}
\end{table*} 
\begin{table*}[h]
 \vspace{-0.15cm}
    \setlength{\tabcolsep}{4.5pt}
    \centering
    \caption{TSC task on 92 UCR datasets.
    The average accuracy, F1 score, and precision on 92 UCR datasets are reported. }
    \vspace{-0.3cm}
    \label{tab:metric_TSC}
    {\footnotesize
    \begin{tabular}{c|c|cccc|cccc}
        \hline
        \multirow{2}{*}{\shortstack{}} &\multirow{2}{*}{Methods} &  \multicolumn{4}{c|}{Overall accuracy}  &  \multicolumn{4}{c}{Overall F1 score}\\
         & & Average Value & Ours Wins & Baseline Wins &Tie & Average Value & Ours Wins & Baseline Wins &Tie  \\ 
         \midrule[0.5pt]
         \multirow{6}{*}{\shortstack{92 UCR \\ datasets \\(only TS)}} &FCNet~\cite{wang2017time} &77.7$\pm$0.0009&\textbf{61.0$\pm$1.25}&14.0$\pm$3.3&17.0$\pm$3.74&77.2$\pm$0.0014&\textbf{61.0$\pm$1.89}&14.0$\pm$1.89&17.0$\pm$3.27\\
         
        &ResNet~\cite{wang2017time}&79.3$\pm$0.0021&\textbf{57.0$\pm$1.41}&17.0$\pm$1.7&18.0$\pm$1.25&78.8$\pm$0.0023&\textbf{57.0$\pm$1.25}&17.0$\pm$0.94&17.0$\pm$2.05\\
        &InceptionTime~\cite{ismail2020inceptiontime} &85.0$\pm$0.0003&\textbf{41.0$\pm$0.47}&26.0$\pm$1.63&25.0$\pm$1.7&84.7$\pm$0.0003&\textbf{42.0$\pm$0.82}&28.0$\pm$1.7&22.0$\pm$0.94\\

         &OSCNN~\cite{tang2020omni} &84.8$\pm$0.0011&\textbf{43.0$\pm$1.41}&19.0$\pm$2.62&30.0$\pm$3.4&84.5$\pm$0.0013&\textbf{42.0$\pm$2.87}&18.0$\pm$1.89&32.0$\pm$3.56\\

    &FormerTime~\cite{cheng2023formertime} &73.5$\pm$0.0024&\textbf{74.0$\pm$1.25}&6.0$\pm$1.41&12.0$\pm$1.7&72.7$\pm$0.0034&\textbf{75.0$\pm$0.82}&6.0$\pm$1.25&11.0$\pm$0.94\\
    &MultiRocket~\cite{41tan2022multirocket}&84.5$\pm$0.0006&\textbf{50.0$\pm$2.45}&22.0$\pm$1.7&20.0$\pm$0.94&84.2$\pm$0.0006&\textbf{49.0$\pm$2.05}&23.0$\pm$2.45&20.0$\pm$0.47\\
    &Ours &\textbf{85.8$\pm$0.0002}&-&-&-&\textbf{85.4$\pm$0.0002}&-&-&-\\
    \hline
    \hline
    \multirow{6}{*}{\shortstack{92 UCR \\ datasets \\(Fusing TS \\and Image)}}
    &MLP Fusion &85.271$\pm$0.0007&\textbf{43.0$\pm$2.05}&23.0$\pm$1.7&26.0$\pm$0.82&84.932$\pm$0.0009&\textbf{45.0$\pm$3.3}&22.0$\pm$3.56&25.0$\pm$1.89\\
    &CNN Fusion &85.542$\pm$0.0011&\textbf{41.0$\pm$1.25}&22.0$\pm$1.7&28.0$\pm$2.49&85.157$\pm$0.0016&\textbf{42.0$\pm$2.83}&21.0$\pm$2.36&29.0$\pm$4.5\\
    &WideDeep Fusion~\cite{cheng2016wide} &85.523$\pm$0.0005&\textbf{46.0$\pm$2.36}&16.0$\pm$1.25&29.0$\pm$2.87&85.191$\pm$0.0004&\textbf{47.0$\pm$1.41}&18.0$\pm$0.0&27.0$\pm$1.41\\
    &Transformer Fusion~\cite{vaswani2017attention} &85.678$\pm$0.0009&\textbf{42.0$\pm$2.16}&21.0$\pm$2.87&29.0$\pm$3.09&85.337$\pm$0.0008&\textbf{42.0$\pm$2.05}&22.0$\pm$2.49&28.0$\pm$2.16\\

    &Adaptive Fusion~\cite{wang2022adaptive} &85.13$\pm$0.0006&\textbf{49.0$\pm$2.16}&16.0$\pm$0.47&27.0$\pm$1.7&84.758$\pm$0.0008&\textbf{48.0$\pm$2.62}&18.0$\pm$1.7&26.0$\pm$2.45\\

    &Ours &\textbf{86.251$\pm$0.0008}&-&-&-
    &\textbf{85.942$\pm$0.0009}&-&-&-\\
        \hline
    \end{tabular}}
\vspace{-0.2cm}
\end{table*}
\vspace{-0.2cm}
\subsection{Results on strain
gauge status recognition}
\textbf{Only TS input.} The results on four groups of offline experimental data in Table~\ref{tab:metric_TSC_SGS} show that our method achieves the best performance. 
For instance, our average accuracy on four groups is 89.91, while baselines (top down in the table) are 88.99, 89.07, 89.51, 89.19, 88.55, and 87.0. 
This indicates that our hypergraph interaction network can better learn high-order relationships between local features to capture global features.

\textbf{Fusing TS, Image, and Expert features.} 
For fusing different features, our average accuracy on four groups is 95.12, while baselines (top down in the table) are 90.98, 93.09, 90.55, 92.72, and 92.66. 
This indicates our hypergraph interaction fusion can better realize feature alignment and re-establish semantic consistency, thereby enhancing representations and improving TSC accuracy through their complementarity. 

It is worth noting that in Experimental Group 3, our method outperforms baselines by a large margin, \textbf{indicating that our approach has better generalization across different experimental datasets.} 
We attribute this to the benefits of high-order interactions in hypergraphs, which can quickly establish associations between unseen data and trained data, thereby achieving better generalization. 
This key advantage means that our method does not require repeated fine-tuning during online deployment.
 \vspace{-0.1cm}
\subsection{Results on UCR datasets}
On TSC experiments of 92 UCR datasets, we observe similar conclusions 
obtained in the scenario of SGS recognition. 
As shown in Table~\ref{tab:metric_TSC}, our method achieves both the best overall accuracy on the settings of single TS input and fusing multiple inputs of TS and image. 
This indicates that our method shows generality.

 \vspace{-0.1cm}
\subsection{Online testing}
In this section, we train top-3 best competitors using the four groups of offline data and deploy them to the web server. 
They are directly used for SGS recognition of aircraft wings in three online experiments. 
Table~\ref{tab:metric_online} shows the average accuracy of our method is 98.17, while for best competitors are 95.03 (CNN fusion), 96.61 (Transformer fusion), and 93.86 (Adaptive fusion).
Our method achieves the best performance, indicating better generalization for new data.

 \begin{table*}[h]
    \setlength{\tabcolsep}{3.5pt}
    \centering
    \vspace{-0.1cm}
    \caption{SGS recognition on three online test experiments based on the web server. The trained model is directly deployed on the web server to obtain prediction results, hence there are no standard deviations reported.
 }
    \vspace{-0.25cm}
    \label{tab:metric_online}
    {\footnotesize
    \begin{tabular}{c|ccc|ccc|ccc|ccc}
        \hline
        \multirow{2}{*}{\shortstack{}}  &  \multicolumn{3}{c|}{Ours}  &  \multicolumn{3}{c}{CNN fusion} &  \multicolumn{3}{c}{Transformer fusion}&  \multicolumn{3}{c}{Adaptive fusion}\\
         &  Accuracy & F1 Score &Precision &Accuracy &F1 Score &Precision &Accuracy &F1 Score &Precision&Accuracy &F1 Score &Precision\\ 

         \midrule[0.5pt]
    \multirow{1}{*}{\shortstack{Online Test 1 (500 samples)}} &\textbf{99.8}&\textbf{99.8}&\textbf{99.8}&97.45&97.45&97.5&98.5&98.5&98.53&96.45&96.3&96.35\\
             \midrule[0.5pt]
    \multirow{1}{*}{\shortstack{Online Test 2 (500 samples)}} &\textbf{98.9}&\textbf{98.9}&\textbf{98.93}&94.5&94.53&94.8&96.95&96.98&97.13&92.45&92.4&92.63\\
             \midrule[0.5pt]
    \multirow{1}{*}{\shortstack{Online Test 3 (5444 samples)}} &\textbf{95.8}&\textbf{95.8}&\textbf{95.88}&93.13&92.98&93.23&94.38&94.4&94.55&92.68&92.65&92.75\\
      \midrule[0.5pt]
    \multirow{1}{*}{\shortstack{Average}} &\textbf{98.17}&\textbf{98.17}&\textbf{98.2}&95.03&94.99&95.18&96.61&96.63&96.74&93.86&93.78&93.91\\

        \hline
    \end{tabular}}
\vspace{-0.1cm}
\end{table*}

\subsection{Ablation study}
\subsubsection{\textbf{Ablation of each component}} 
In this section, we remove various components in the method to observe changes in model accuracy. 
Results in Table~\ref{tab:ablation_tots} indicate that the model's accuracy shows a decrease while different components are removed, which verifies their effectiveness. 
Specifically, w/o Dynamic hyperedges means we use hyperedges in the initially learned hypergraph for information propagation, instead of using the new one from the newly learned hypergraph. 
w/o RWHC means we use KNN to select nodes for constructing hyperedges. 
w self-attention means we use it to replace the hypergraph attention (Figure~\ref{fig:frame_work}(d)). 
w/o Key embedding means we use Eq.~\ref{equ:attn1} with new parameters to obtain Key for calculating $Att$ in Eq.~\ref{equ:attn2}. 
\textbf{Moreover, we compare the nodes’ proportion in hyperedges with different constructions (Random-walk VS KNN) in Figure~\ref{fig:hyperG_res} of the appendix due to limited space.} 
Our random-walk based hyperedges can fully consider the correlation among all three types of nodes, enhancing representations.

\begin{figure}[h]
\centerline{\includegraphics[width=1\linewidth]{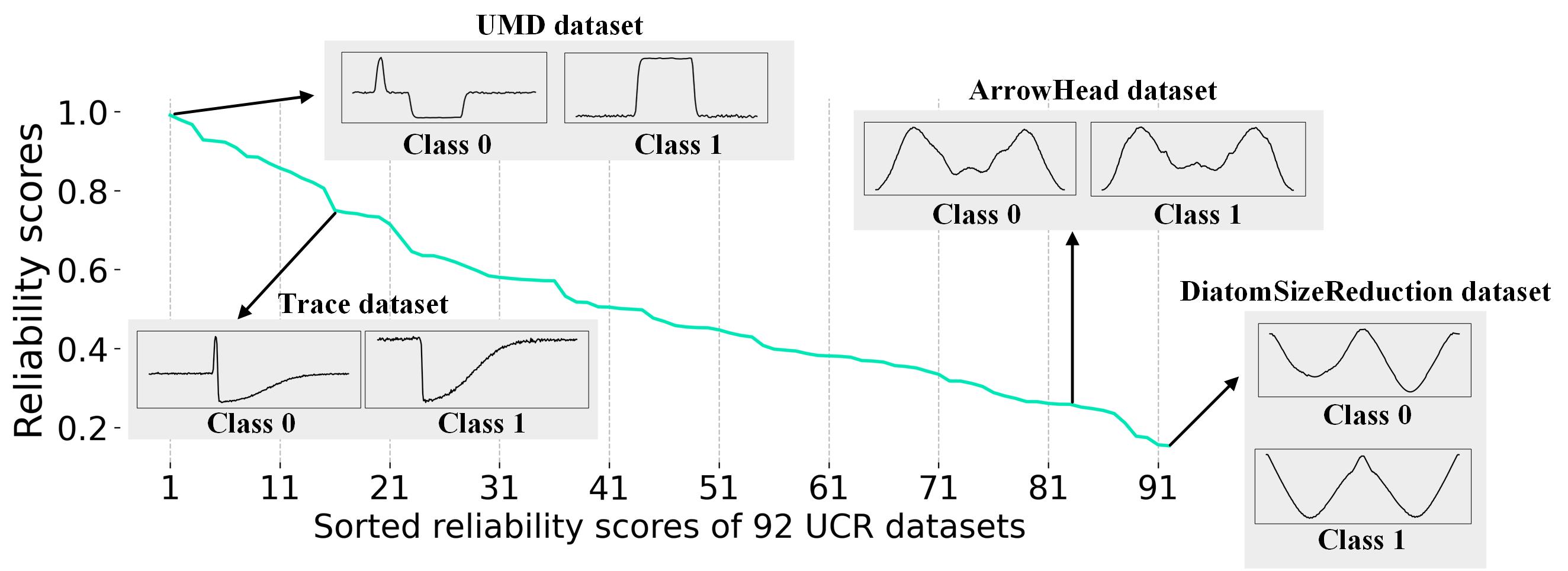} }
\vspace{-0.2cm}
\caption{Sorted average DRA scores on different datasets. }
\vspace{-0.1cm}
\label{fig:relaibility_scores}
\end{figure}

\vspace{-0.1cm}
\subsubsection{\textbf{Discussion of Data Reliability-aware
Attention (DRA) scores}} 
We average all image samples' DRA scores for the dataset. 
Figure~\ref{fig:relaibility_scores} visualizes the samples from different datasets corresponding to the average DRA scores. 
The results indicate that for datasets with higher DRA scores, the profiles of different samples generally have a more distinct discriminability, while for those with lower scores, they are more similar and easily confused. 
Our method successfully assigns higher scores to datasets that are suitable for utilizing profile features to enhance representations, thereby improving accuracy. 
This further shows the DRA mechanism's effectiveness.

 \begin{table} [bt] 
    \centering
    \caption{Ablation Study. DRA denotes Data Reliability-aware Attention. RWHC denotes Random Walk Hypergraph Construction. We report average accuracy on 92 UCR datasets.}
    \vspace{-0.3cm}
    \label{tab:ablation_tots}
    { \small
    \begin{tabular}{c|c|c}
        \hline
        \multirow{1}{*}{\shortstack{Datasets}}  &  \multicolumn{1}{c|}{\shortstack{Average Accuracy}} &  \multicolumn{1}{c}{\shortstack{Average F1 Score}}\\
         \midrule[0.5pt]
         {\shortstack{Ours}}  &\textbf{86.251$\pm$0.0008}&\textbf{85.942$\pm$0.0009} \\
         \midrule[0.5pt]
        {\shortstack{w/o Dynamic hyperedges }}  &85.946$\pm$0.0013&85.593$\pm$0.0013 \\
        \midrule[0.5pt]
        {\shortstack{w/o RWHC}}  &85.743$\pm$0.0009&85.4$\pm$0.001  \\
        \midrule[0.5pt]

        {\shortstack{w/o DRA}}  &85.592$\pm$0.0006&85.241$\pm$0.0006  \\
        \midrule[0.5pt]
        
        {\shortstack{w/o Hypergraph attention}}  &85.778$\pm$0.001&85.442$\pm$0.0012 \\
        \midrule[0.5pt]
        {\shortstack{w self-attention}}  &85.634$\pm$0.0016&85.29$\pm$0.0015 \\

        \midrule[0.5pt]

        {\shortstack{w/o Key embedding}}  &86.113$\pm$0.0002&85.825$\pm$0.0005 \\
        \midrule[0.5pt]

    \end{tabular}
    } 
\vspace{-0.3cm}
\end{table}

\subsubsection{\textbf{Effectiveness of feature redundancy filtering (FRF in Figure~\ref{fig:frame_work}(b))}}
We propose to use learnable binary masks to filter the redundant information in the features extracted by CNNs, allowing the model to better utilize effective feature parts. 
Results in appendix Table~\ref{tab:FRF_tab} show that the model with FRF has more wins on 92 UCR datasets. 
Moreover, we visualize the interest regions of the 1DCNNs through Grad-CAM~\cite{selvaraju2017grad} with the setting of w FRF and w/o FRF in appendix Figure~\ref{fig:FRF_heat_map}.
The results show that w FRF not only retains attention to the original areas that are captured by w/o FRF, but also pays more attention to other important areas, such as Tag 2, Tag 5, and Tag 4 in the appendix Figure~\ref{fig:FRF_heat_map}. 
The above analysis shows the effectiveness of the FRF design.

\subsubsection{\textbf{Effectiveness of global feature engineering}}
To enhance the representation of TS samples, we additionally constructed image data and expert features. 
Table~\ref{tab:ablation_image_exp} shows that removing image data or expert features (w/o Expertise) can lead to a decrease in accuracy, indicating that the profile features in the images and expert features from expertise (e.g., polynomial's coefficient to measure the curve's curvature) can indeed help better represent the samples, thereby improving TSC accuracy. 
\textbf{Moreover, Table~\ref{tab:only_image_exp} of appendix shows accuracy with only single input (only $X_{ts}$, only $X_{img}$, and only $X_{exp}$ respectively), both suffering from significantly decrease in accuracy. Only fusing all of them can achieve the best accuracy, further indicating the effectiveness of our approach.}

 \begin{table} [bt] 
    \centering
    \caption{Exploring the effectiveness of global feature engineering, e.g., image data $X_{img}$ and expert features $X_{exp}$.}
    \vspace{-0.3cm}
    \label{tab:ablation_image_exp}
    { \small
    \begin{tabular}{c|c|c|c}
        \hline
        \multirow{1}{*}{\shortstack{Datasets}}  &  \multicolumn{1}{c|}{\shortstack{Ours}} &  \multicolumn{1}{c|}{\shortstack{w/o Image}}&  \multicolumn{1}{c}{\shortstack{w/o Expertise}}\\

         \midrule[0.5pt]
        {\shortstack{Strain gauge G1 }}&\textbf{92.875$\pm$0.0014}&92.333$\pm$0.0043&88.225$\pm$0.0004 \\
        \midrule[0.5pt]
        
        {\shortstack{Strain gauge G2}} &\textbf{97.517$\pm$0.0012}&97.408$\pm$0.0004&93.25$\pm$0.002 \\
        \midrule[0.5pt]
        {\shortstack{Strain gauge G3}} &\textbf{95.1$\pm$0.002}&94.792$\pm$0.002&89.742$\pm$0.0029\\
        \midrule[0.5pt]
        {\shortstack{Strain gauge G4}}&\textbf{94.967$\pm$0.0017}&94.775$\pm$0.0007&88.392$\pm$0.0013 \\
        \midrule[0.5pt]
         {\shortstack{92 UCR Datasets}}  &\textbf{86.251$\pm$0.0008} &85.8$\pm$0.0002&- \\
        \midrule[0.5pt]
        
    \end{tabular}
    } 
\vspace{-0.5cm}
\end{table}

\section{Conclusion}
This paper attempts to introduce global features for strain gauge status recognition in the industrial scenario to address the issue that existing local features (extracted by 1D-CNNs) are insufficient to express time series samples. 
Meanwhile, we propose a hypergraph interaction fusion framework, which can not only learn global features based on local features but also realize feature alignment and re-establish semantic consistency among different global features. 
Our approach successfully enhances the sample's representation and improves TSC accuracy on both real industrial datasets and public UCR datasets, showing better generalization for unseen data.

\bibliographystyle{ACM-Reference-Format}
\bibliography{sample-base}

\appendix
\begin{table*}[bt]
\hspace{0.2cm}
\begin{minipage}{0.35\textwidth}
\setlength{\tabcolsep}{3.5pt}
\caption{Wins comparison on 92 UCR datasets of using Feature Redundancy Filtering (Ours w FRF) and not using it (Ours w/o FRF). The case study of FRF on the class activation map is shown in Figure~\ref{fig:FRF_heat_map}.}
\vspace{-0.2cm}
\label{tab:FRF_tab}
   \begin{tabular}{c|c|c}
        \hline
        \multirow{1}{*}{\shortstack{Datasets}}  &  \multicolumn{1}{c|}{\shortstack{w FRF}} &  \multicolumn{1}{c}{\shortstack{w/o FRF}}\\

        \midrule[0.5pt]
         {\shortstack{Wins on 92 UCR datasets}}  &\textbf{34$\pm$0.5} &24$\pm$1.5  \\
         
        \midrule[0.5pt]  
      {\shortstack{Tie on 92 UCR datasets}}
       & \multicolumn{2}{c}{\shortstack{34$\pm$2.0}} \\
        \midrule[0.5pt]

    \end{tabular}
\end{minipage}%
\hspace{0.15cm}
\begin{minipage}{0.6\textwidth}
  \centering
  \includegraphics[width=0.9\linewidth]{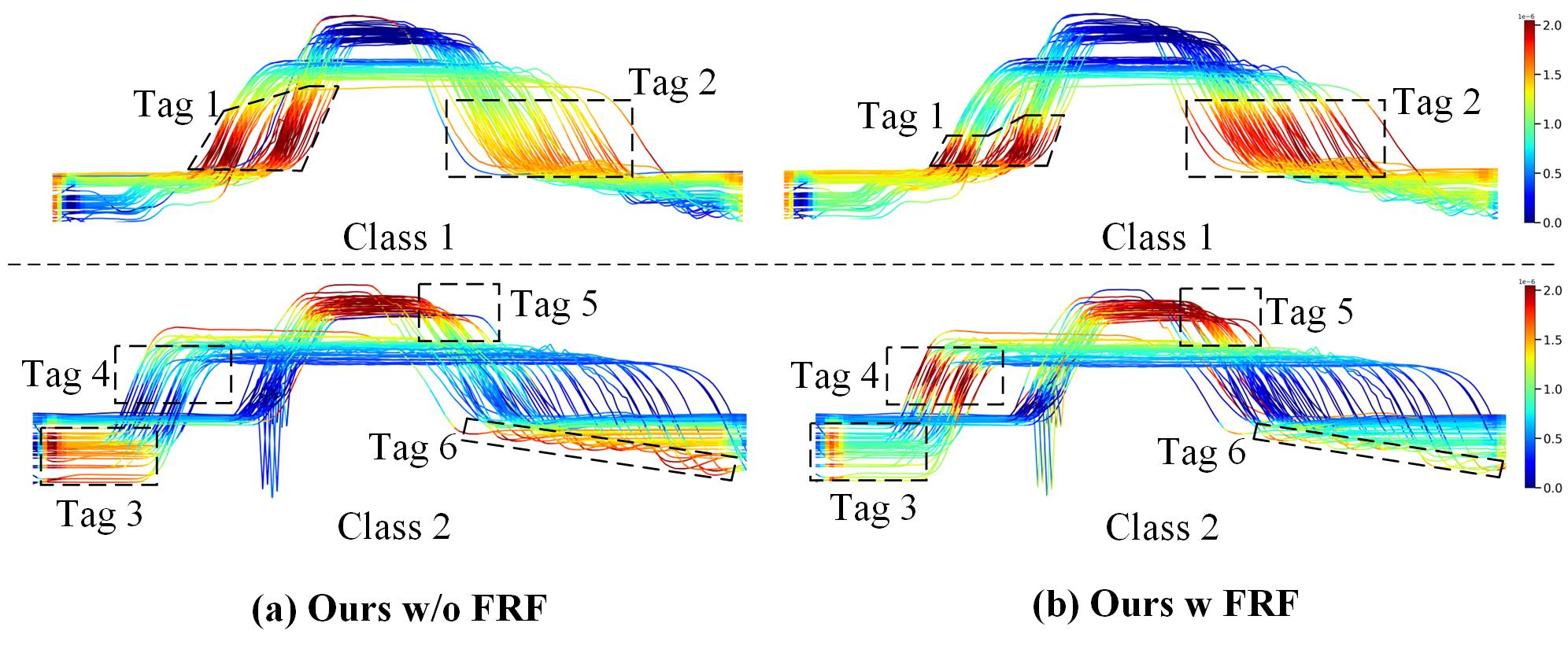}
  \vspace{-0.5cm}
  \captionof{figure}{Class activation map on the GunPoint dataset.}
  \label{fig:FRF_heat_map}
\end{minipage}%
\vspace{-0.1cm}
\end{table*}
\section{Appendix}

\subsection{More main results and ablation study}
\subsubsection{\textbf{Extra results of only using image and expert features}}
Results in Table~\ref{tab:only_image_exp} show that only time series $X_{ts}$, only image data $X_{img}$ and only expert features $X_{exp}$ as model input can lead to significant decrease
in accuracy. 

This indicates that the ability of a single input of time series, images, or expert features for representing TS samples has limitations. 
\textbf{However, when they interact with each other on our designed hypergraph network (``Fusing all'' in Table~\ref{tab:only_image_exp}), a significant increase in accuracy is observed.} 
This demonstrates that our proposed high-order hypergraph interactions can effectively utilize the correlations between cross-type features to enhance representations, thereby improving classification accuracy.

\vspace{-0.1cm}
\begin{table} [bt] 
  \setlength{\tabcolsep}{3.5pt}
    \centering
    \caption{Accuracy of only TS strain data  $X_{ts}$, only image $X_{img}$ and only expert features $X_{exp}$ as model input respectively on four strain gauge experiment groups.}
    \vspace{-0.3cm}
    \label{tab:only_image_exp}
    { \small
    \begin{tabular}{c|c|c|c|c}
        \hline
        \multirow{1}{*}{\shortstack{}}  &  \multicolumn{1}{c|}{\shortstack{Fusing all (Ours)}} &  \multicolumn{1}{c|}{\shortstack{Only TS}}&  \multicolumn{1}{c|}{\shortstack{Only Image}}&  \multicolumn{1}{c}{\shortstack{Only Expertise}}\\

         \midrule[1pt]
        {\shortstack{G1 }}&\textbf{92.875$\pm$0.0014}&88.017$\pm$0.0019&75.183$\pm$0.0084&87.508$\pm$0.0006\\
        \midrule[1pt]
        
        {\shortstack{G2}} &\textbf{97.517$\pm$0.0012}&93.517$\pm$0.002&89.383$\pm$0.0489&91.008$\pm$0.0014 \\
        \midrule[1pt]
        {\shortstack{G3}} &\textbf{95.1$\pm$0.002}&89.642$\pm$0.0023&85.975$\pm$0.0608&81.1$\pm$0.037\\
        \midrule[1pt]
        {\shortstack{G4}}&\textbf{94.967$\pm$0.0017}&88.45$\pm$0.0011&84.15$\pm$0.0634&77.717$\pm$0.0299 \\
        \midrule[1pt]
        
    \end{tabular}
    } 
\vspace{-0.3cm}
\end{table}

\begin{figure}[bt]
\centerline{\includegraphics[width=1\linewidth]{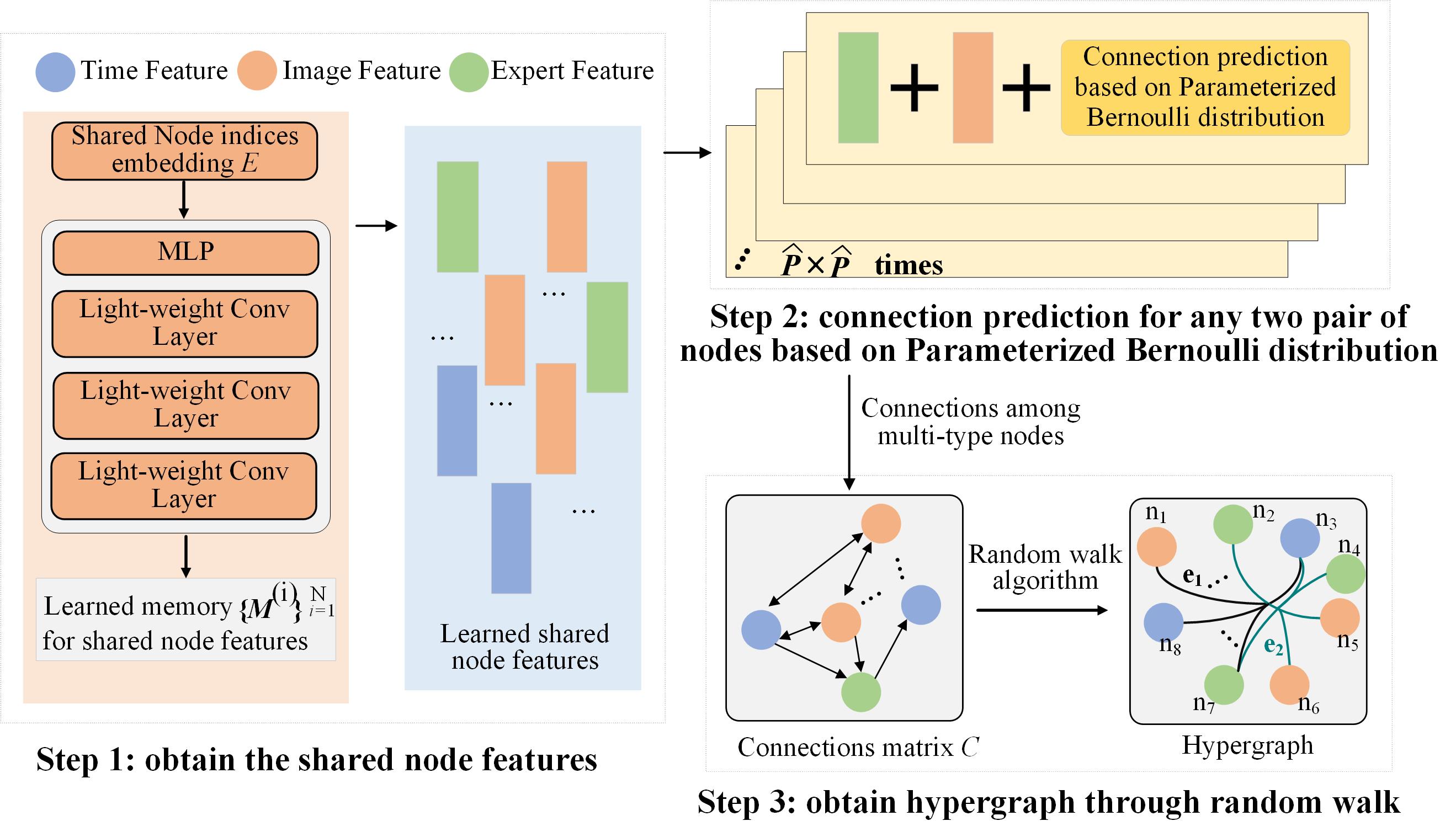} }
\vspace{-0.4cm}
\caption{Illustration of Shared Hypergraph Extractor of the proposed framework in Figure~\ref{fig:frame_work}. }
\vspace{-0.2cm}
\label{fig:hypg_learn}
\end{figure}

\vspace{-0.1cm}
\subsubsection{\textbf{Illustration of hypergraph
construction}}
Here we further illustrate the hypergraph construction in Figure~\ref{fig:hypg_learn} with three steps. 
\textbf{In step 1}, we learn the shared node features. \textbf{In step 2}, we concatenate the features of any two nodes and feed them into the parameterized Bernoulli distribution to predict whether there is a correlation (connection) between them. 
There are a total of $\hat{P}
\times \hat{P}$ connection predictions.
\textbf{In step 3}, all connections between nodes together form a graph, and finally, we perform random walks on the graph to construct hyperedges, forming the hypergraph.

\subsubsection{\textbf{Nodes' proportion in hyperedges with different constructions}}
\begin{figure}[bt]
\centerline{\includegraphics[width=1.1\linewidth]{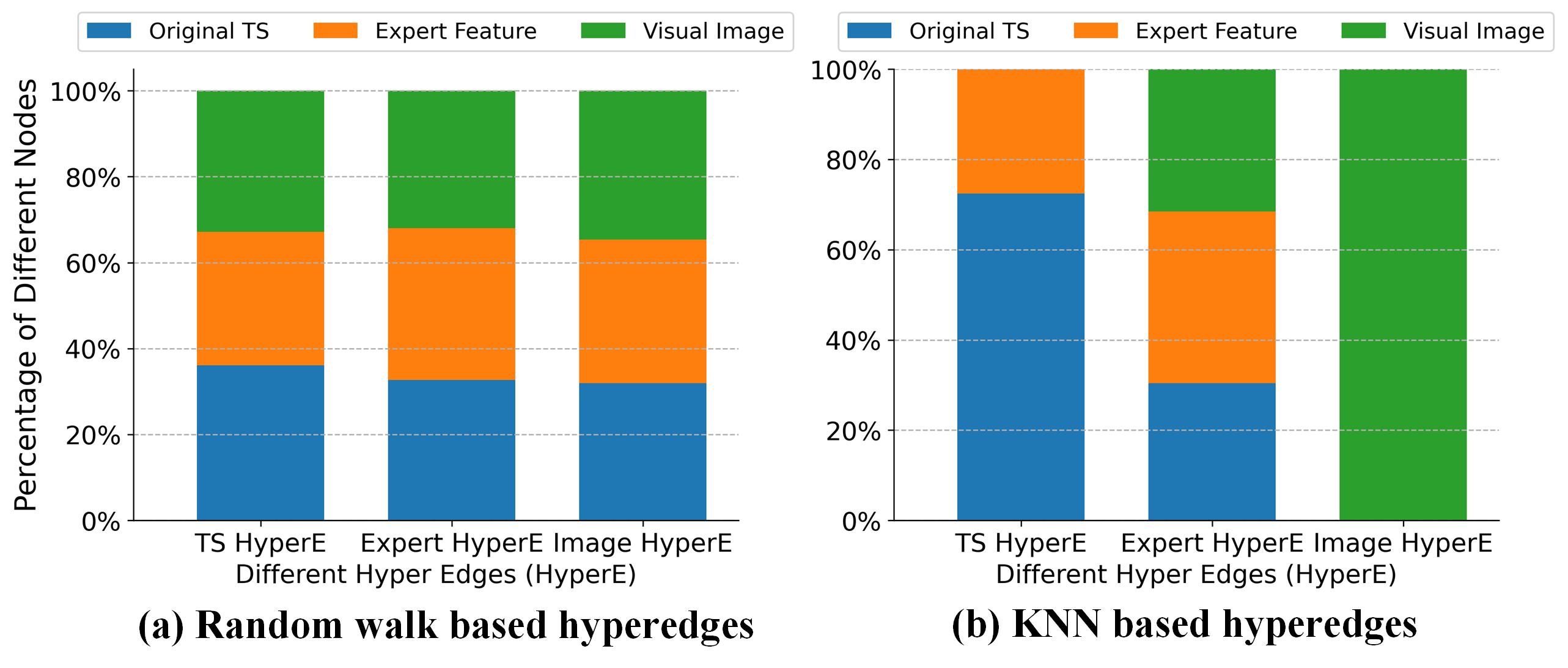} }
\vspace{-0.2cm}
\caption{The proportion of different types of nodes within different hyperedges. The left figure shows statistical results on random-walk (ours) based hyperedges while the right is results on KNN-based hyperedges. }
\vspace{-0.2cm}
\label{fig:hyperG_res}
\end{figure}

In the main paper, we analyze that KNN-based hyperedges construction may not be suitable in our scenario with multi-type node features. 
we have demonstrated this in Figure~\ref{fig:hyperG_res}. 

Specifically, we calculate the proportion of different types of nodes within different hyperedges, and the results show that the hyperedge construction of KNN can't effectively include cross-type nodes. For example, as shown in Figure~\ref{fig:hyperG_res}(b), the $Z_{ts}$'s hyperedges only include nodes in $Z_{ts}$ and $Z_{exp}$. 
The $Z_{img}$'s hyperedges only includes nodes of $Z_{img}$ itself. 
This limits the utilization of inherent correlations of cross-type nodes to enhance representations. 
\textbf{In contrast, our random-walk based hyperedges in Figure~\ref{fig:hyperG_res}(a) all fully consider the nodes of three types.}

\subsubsection{\textbf{Conventional local-feature based TSC}}
Conventional method relies on manually designing local feature extraction algorithms, e.g., shapelets methods~\cite{DBLP:journals/datamine/YeK11,DBLP:journals/datamine/HillsLBMB14} and interval statistic features~\cite{DBLP:journals/kbs/RodriguezAM05,DBLP:journals/datamine/LubbaSKSFJ19} (e.g., mean and slope). 
These methods inevitably incur expensive computations in producing discriminative local features~\cite{middlehurst2024bake}, especially when expanding for large-scale data. 
In contrast, the MultiRocket~\cite{41tan2022multirocket}, which extracts local features using randomly parameterized convolution kernels, offers higher efficiency and competitive accuracy.

\end{document}